\documentclass{article}

\usepackage{arxiv}

\usepackage[utf8]{inputenc} 
\usepackage[T1]{fontenc}    
\usepackage{hyperref}       
\usepackage{url}            
\usepackage{booktabs}       
\usepackage{amsfonts}       
\usepackage{nicefrac}       
\usepackage{microtype}      
\usepackage{lipsum}
\usepackage{graphicx}

\usepackage{amsmath}
\usepackage{tabularx}

\title{Bond Default Prediction with Text Embeddings, Undersampling and Deep Learning}

\author{Luke Jordan \\
  MIT GOV/LAB \\
  Cambridge, MA \\
  \texttt{lukej@mit.edu}
}

\begin{document}

\maketitle

\begin{abstract}
    The special and important problems of default prediction for municipal bonds are addressed using a combination of text embeddings from a pre-trained transformer network, a fully connected neural network, and synthetic oversampling. The combination of these techniques provides significant improvement in performance over human estimates, linear models, and boosted ensemble models, on data with extreme imbalance. Less than 0.2\% of municipal bonds default, but our technique predicts 9 out of 10 defaults at the time of issue, without using bond ratings, at a cost of false positives on less than 0.1\% non-defaulting bonds. The results hold the promise of reducing the cost of capital for local public goods, which are vital for society, and bring techniques previously used in personal credit and public equities (or national fixed income), as well as the current generation of embedding techniques, to sub-sovereign credit decisions.
\end{abstract}


\section{Introduction}

Municipal bonds in the United States are a \$4 trillion market. They fund schools, water works, transport systems and much else. Only 0.1\% of municipal bonds default, but under Basel III regulations, they carry risk weightings equivalent to commercial banks and credit unions, for bonds that are backed by general taxes (``general obligations''), or residential mortgages, for bonds backed by more specific municipal income streams (``revenue bonds''). Municipal bonds are rated by the large credit rating agencies, at cost to the issuers. While municipal bonds are traded, their most important price is their yield at issue-the amount that municipalities must pay on the bonds. The setting of such prices is highly opaque, even post-issue. Researchers have found that market participants incur substantial costs just to probe the quality of issues \cite{seru2010learning}. On the other hand, their public nature means that municipal bonds have a high degree of transparency, with large, consistent and easily available datasets stretching back many decades.

These characteristics make municipal bonds fertile ground for the application of machine learning and artificial intelligence (AI). While AI has been used extensively for default prediction, such uses have been primarily for personal credit, raising many ethical and regulatory concerns, from bias to interpretability, and the social welfare benefits of such results are highly contestable. While the risk of bias remains present for municipal bonds, for example learned discrimination against low-income or minority municipalities, it is likely to be significantly simpler to monitor and mitigate. Conversely, improved default prediction independent of issued ratings might at least reduce the direct cost of issuance, and at best provide a strong motivation for reducing the Basel III risk-weighting for municipal bonds, unlocking a large amount of additional capital for local public goods. If the inference functions of such models were made easily accessible, they could help ordinary people understand the risks and potential pricing of the debt their cities issue (or propose to issue). Finally, municipal bonds are a special category of what are called impact bonds, indeed the single largest category of impact bonds. The plentiful, clean and consistent data available for municipal bonds also make them a strong starting point for developing models that might in the future be transferred to other categories of impact bonds, or local government credit in many other jurisdictions.

With that motivation, we use a combination of machine learning techniques to estimate municipal bond defaults using an intentionally limited set of features. We do not include credit ratings or time series data on the municipalities issuing the bonds. Instead, we use only easily and publicly available data on bond purposes, duration, and location and widely observed, national macroeconomic data. We do, however, enrich the data by using the latest generation of pretrained transformer-based natural language processing models to generate embeddings of the project descriptions published for each bond. We use synthetic oversampling on the dataset as a whole, and weighted random sampling in selecting batches during training, to handle extreme imbalance. Using the combination of these methods on a large dataset, we are able to achieve significant advances over linear models and implied human performance. 

As one import clarification, we do not seek to optimize the \emph{trading} of municipal bonds, and likewise do not seek to improve the post-issue pricing of such bonds. The secondary market for municipal bonds plays an important role in the viability, liquidity and depth of the municipal bond market, but it is not our focus here. Our focus is on the decision making involved in the initialissuance and purchasing of the debt, that is, on the primary processes by which real-world projects are built through credit allocation decisions. We do however see clear directions for future research in the extension of these methods to post-issue trading and market making.

\section{Related Work}

Credit default prediction has been extensively studied, including the problem of imbalanced datasets, in the context of personal credit (such as credit cards, other unsecured lending, mortgages, and so forth). \cite{shen2021new} recently used synthetic oversampling, with the introduction of a Mahalanobis distance based separation boundary in sample generation to reduce noise, combined with a long-short term memory (LSTM) network as base learner in an adaptive boosting algorithm to achieve state of the results on UCI personal credit datasets. \cite{hamori2018ensemble} conducts a review of prior literature and of ensemble and neural network methods, making some reference to results on corporate credits but performing their principal evaluation on personal credit data. While on a somewhat different domain, we have drawn on the techniques for handling extremely imbalanced datasets developed in this literature. 

For corporate credit, \cite{moscatelli2020corporate} studied corporate defaults in Italy using gradient-boosted ensemble methods, finding that they significantly outperformed traditional linear models when only public data was available, and adapt more quickly to changes in the macroeconomy, and draw out regulatory implications from those results. \cite{kim2020corporate} provides a review of earlier corporate default methods, noting the challenges of imbalanced datasets, and finding promising results using ensemble methods and support vector machines. \cite{zhou2013performance} earlier reviewed the effects of sampling methods on corporate default predictions.

Public debt has been studied predominantly for post-issue trading and pricing. \cite{raman2018municipal} uses hierarchical structural models to estimate post-issue secondary market prices for municipal bonds on a given day using only prior trading data, as well as the sector and credit rating of a bond. \cite{bond-risk-premia} predicts excess Treasury bond returns using random trees and neural networks. They show large gains compared to the state of the art in algorithmic trading, with neural network models gaining from the introduction of macroeconomic variables. \cite{gotze2020improving} evaluated pricing for a small (< 600) dataset of ``catastrophe'' bonds, finding that ensemble methods could match the performance of extensively tested and deployed linear models, with lower variance.

\section{Data}

Our data comes from the Mergent Municipal Bond Securities Dataset \cite{mergent}, containing 4 million bonds (maturities) in 445 000 issues, relating to $\approx$60000 projects by US municipalities. Each bond entry contains 94 attributes, including its coupon yield, maturity date, and ratings by the major fixed income rating agencies (S\&P, Moody's, Fitch). Each issue contains data on the project location and has a project description field, containing $28 \pm 15$ words. For macroeconomic data, we use data from the Federal Reserve's FRED service, utilizing a reference safe asset yield (3 month Treasury bills), prior year gross domestic capita (GDP) growth, and inflation rates (CORE CPI).

The dataset is large but on important dimensions is distributed densely and with some skew. The mean spread of yields at issue to Treasury bill yields is $1.25\%$ with a standard deviation of $1.6\%$, but a quarter of bonds are concentrated in a small interval around zero. The data also contains a flag for defaulted maturities, with 4 493 defaults in the dataset, comprising 0.1\% of the maturities (vs \~30\% in the datasets used in \cite{shen2021new}). The most common tokens in the descriptions, as would be expected, relate to financing terms (``revenue'', ``obligation'') and project categories (``school'', ``housing''). The ten most common tokens represent half of all words in the descriptions, but the distribution has a long tail, with $495$ tokens occurring in at least 100 issue descriptions and another ten thousand occurring in fewer than $100$ issues.

\section{Methods}

We train a variety of linear and non-linear supervised learning models using multiple sampling methods to predict whether or not a specific maturity will default. We include the initial coupon offered, but exclude the final yield paid as that encodes average market participant views, and so might leak the label, and will in any case not be available to either an analyst or member of the general public considering a prospective issue. We conduct standard prepossessing of the continuous and categorical attributes, resulting in a somewhat sparse $230$-dimensional feature vector for each bond. 

We then concatenate to each bond an embedding of the project description for which it was issued. The embeddings are produced using a Siamese BERT network \cite{reimers-2019-sentence-bert} recently pretrained for similarity matching, obtained through the HuggingFace project. We do not fine-tune the embeddings, which would be resource intensive, and, given the limited underlying variation and number of text tokens (relative to the demands of large-scale NLP models), would be unlikely to yield significant benefits, and would run the significant risk of creating hidden or explicit overfitting problems. The purpose of the embeddings is to enable the downstream model to utilize the interactions of the descriptions and other features, for which a high-quality similarity matching pre-trained model suffices (as confirmed by the empirical results below). The pretrained network produces embeddings of dimension $d=384$, for a complete input vector in $\mathbb{R}^{614}$. 

A range of models and sampling methods are tested in the experiment: logistic regression, gradient-boosted trees, and several neural network architectures. For each, we test with and without a prior step of utilizing the Synthetic Minority Oversampling Technique (SMOTE) \cite{chawla2002smote}. Specifically, we use the SMOTE-ENC variant recently developed to handle nominal as well as continuous features (SMOTE-ENC) \cite{mukherjee2021smote}. We test further sampling techniques within the training loop for each method, including weighted loss functions. For the neural network models, we use weighted random sampling in the composition of each training batch. We conduct hyperparameter search for each method, and test with and without the project description embeddings. The full set is described in Table~\ref{tab:models}.

\begin{table*}
  \centering
  \caption{Hyperparameter Search-space}
  \label{tab:models}
  \begin{tabular}{p{0.25\linewidth}p{0.65\linewidth}}
    \toprule
    Algorithm & Hyperparameters \\
    \midrule
    Logistic regression & $C \in [0, 1]$ \\
    Gradient boosted trees (XGBoost) & Learning rate $ \in [0.01, 0.1]$, maximum depth $\in [3, 12]$, minimum child weight $\in [1, 10]$, early stopping  \\
    Multi-layer perceptron & Learning rate $\in [0.001, 0.1]$, activation $\in \{\text{'relu'}, 
    \text{'leaky relu'}, \text{'tanh'}\}$, epochs $\in [5, 10, 20]$, batch size $\in [2^{4}, 2^{8}]$,
    dropout $\in \{\text{none}, 0.1, 0.5\}$\\
  \bottomrule
\end{tabular}
\end{table*}

\begin{figure}[ht]
  \centering
  \includegraphics[width=\linewidth]{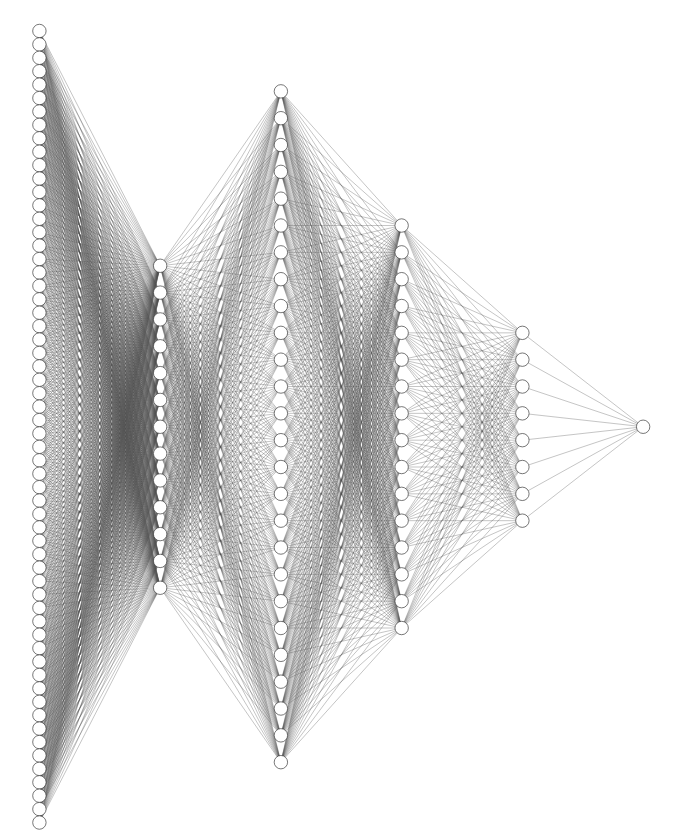}
  \caption{Default Prediction MLP Architecture. Note: For illustration, earlier layers are not drawn precisely to scale}
  \label{fig:mlparch}
\end{figure}

For the neural network models, we relied on moderate sized fully connected networks, or multi-layer perceptrons (MLPs). Given the large but not enormous data available, its limited dimensionality and sparseness, we explored deep and narrow architectures, up to an 8-layer network with an initial layer size of $512$ neurons, but found limited gains. The best performing network had four hidden layers, as drawn in Figure~\ref{fig:mlparch}, with hidden layers sizes $[128, 256, 64, 8]$. The network was trained with dropout ($p=0.1$) and a batch size of $256$. We considered but did not pursue the much more complex models adapting transformer architectures for tabular data \cite{huang2020tabtransformer}. We did not do so since our data is clean with little noise in its key features, required no semi-supervised learning, and already outperformed tree-based methods by a significant margin, and hence did not satisfy the primary criteria given for when such models are worth their significantly increased resource costs. This decision was further supported by the limited gains seen for the deeper MLPs.

For default prediction, given the extreme imbalance in the dataset (<1\% of samples in the minority class) we utilize the area under the precision-recall curve (PR AUC) as our metric, since that focuses attention on the minority class (more so than receiver operating characteristic, or ROC AUC, would). In conducting hyperparameter search we use stratified cross-fold validation, and conduct tests on a hold-out set. For results on our hold-out test set, we utilize both PR AUC and the Kolmogorov–Smirnov two-sample statistic, a common metric in default prediction tasks.

\section{Results}

\begin{figure*}[ht]
  \centering
  \includegraphics[width=\linewidth]{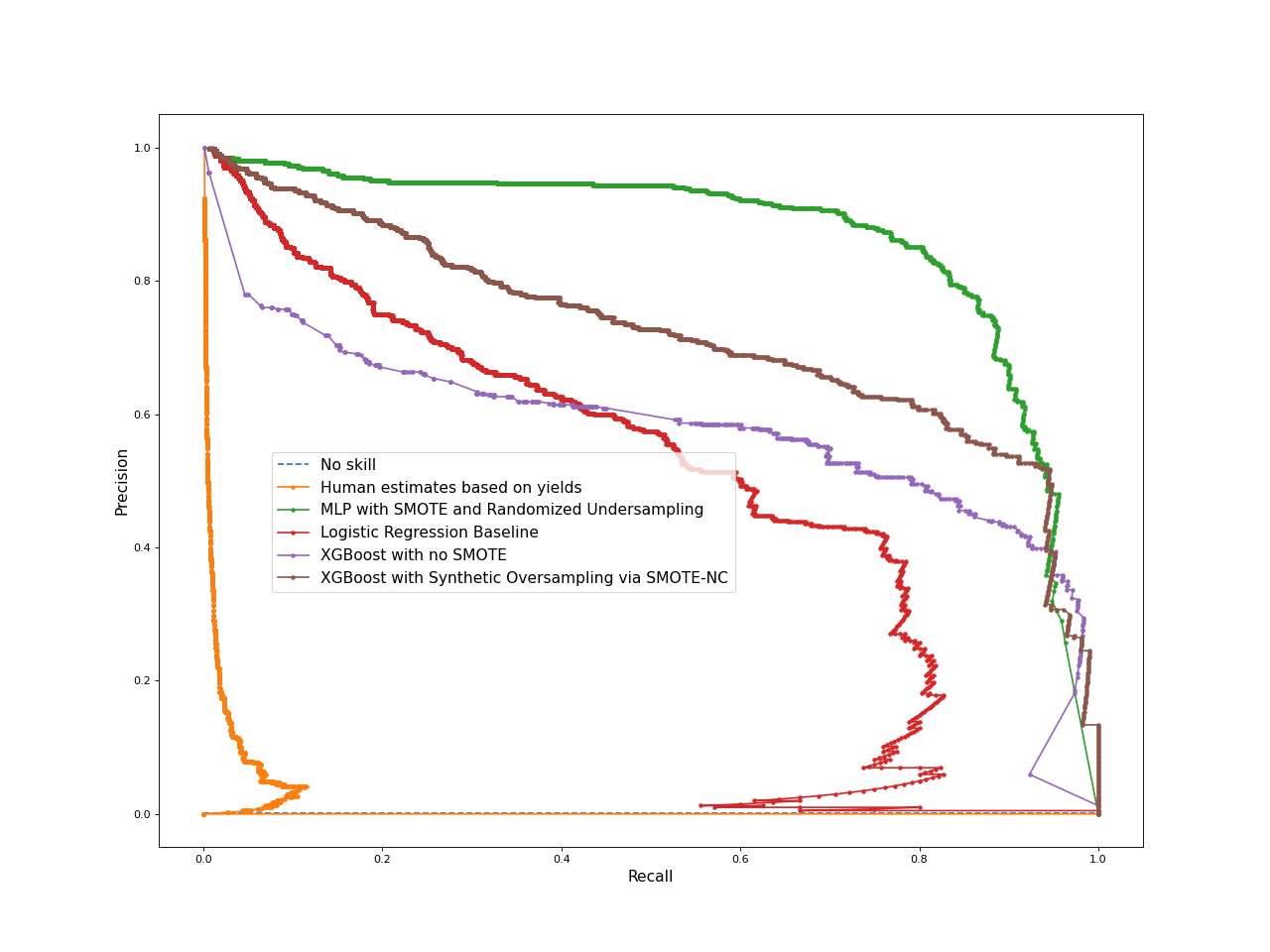}
  \caption{Precision-Recall Curves}
  \label{fig:prcurves}
\end{figure*}

Utilizing each method we obtain the results shown in Table~\ref{tab:results}, with precision-recall curves in Figure~\ref{fig:prcurves}. We report explicitly the false negative and false positive rates (at a decision boundary of $0.5$), since each is a high stakes decision: a false negative representing a significant loss of capital, and a false positive representing a project that should be funded but would not be. 

The non-linear methods demonstrate clear superiority over logistic regression, and the neural network does so against gradient-boosted ensembles. A four-hidden-layer MLP model using moderate ($p=0.1$) dropout and weighted random sampling to balance each batch during training outperforms a gradient-boosted ensemble model with prior synthetic oversampling. On a test set of almost 400,000 bonds with four hundred defaults, the best-performing MLP model misses only 33 defaults and misclassifies only 227 non-defaulters as defaults. 

To place these results in perspective, we have two methods to compare them to human-level performance. One is the ratings at the time of issue, given by the major rating agencies. Since the model is predicting default, it can be compared to the agencies issuing their lowest grade. Note that this test is somewhat biased against the model, since on their own scale the agencies should issue the lowest rating to a maturity they believe will default. However, such a maturity would likely not be able to come to market, and hence would not be observed. Nonetheless, on this evaluation, and taking a rating of "C" and below as indicating a judgement of likely default, the ratings agencies managed to predict at most 97 of 404 defaulting bonds, for a recall rate slightly below the logistic regression baseline, and significantly below the tree-based and deep learning based models. On the other hand, the false positive rate equals the best performing model on that dimension, with only 25 non-defaulting bonds labelled as likely to default. On the other hand, it is precisely on this measure that we would expect selection bias to weight most heavily in the ratings' favour, and, given the size of the test set, is not a meaningful improvement in performance (being a difference of just one in two thousand bonds) against either the deep learning or boosted-tree models.

A second method is to observe the excess yield at-issue provided by the maturity. That excess yield represents the additional risk that market participants, in aggregate, believe that the bond carries of defaulting (net of expectations for some recovery post-default). While a precise calculation of the implied risk of default in any given excess yield is beyond the scope of this paper, combining results from \cite{vrugt2011estimating} and general market rules of thumb, we will assume that a bond with a spread above 10\% is considered by the market a near-certain default. For the sake of comparison, we assume a linear map of spreads between 0\% and 10\% to an implied probability of default, and with that assumption plot the ``humane estimate'' precision-recall curve in Figure~\ref{fig:prcurves}.

\begin{table*}
  \caption{Default Prediction Results on Test Set (N=376598 bonds)}
  \label{tab:results}
    \begin{tabularx}{\linewidth}{lrrrrrr}
    \toprule
    Description &  AUC PR &    KS &  False Positive &  False Negative &  True Positive &  Recall \\
    \midrule
    MLP-SMOTE &   \textbf{0.865} & \textbf{0.967} &             227 &              \textbf{33} &            \textbf{371} &   \textbf{0.918} \\
    Logistic Regression &   0.502 & 0.937 &              50 &             244 &            160 &   0.396 \\
    XGBoost &   0.580 & 0.793 &              \textbf{25} &         226 &            178 &   0.441 \\
    XGBoost-SMOTE &   0.722 & 0.948 &             357 &             103 &            301 &   0.745 \\
    Human estimates based on yields &   0.014 & 0.416 &            3838 &             330 &             74 &   0.183 \\
    \bottomrule
    \end{tabularx}
\end{table*}

\section{Discussion}

\begin{figure*}[ht]
  \centering
  \includegraphics[width=\linewidth]{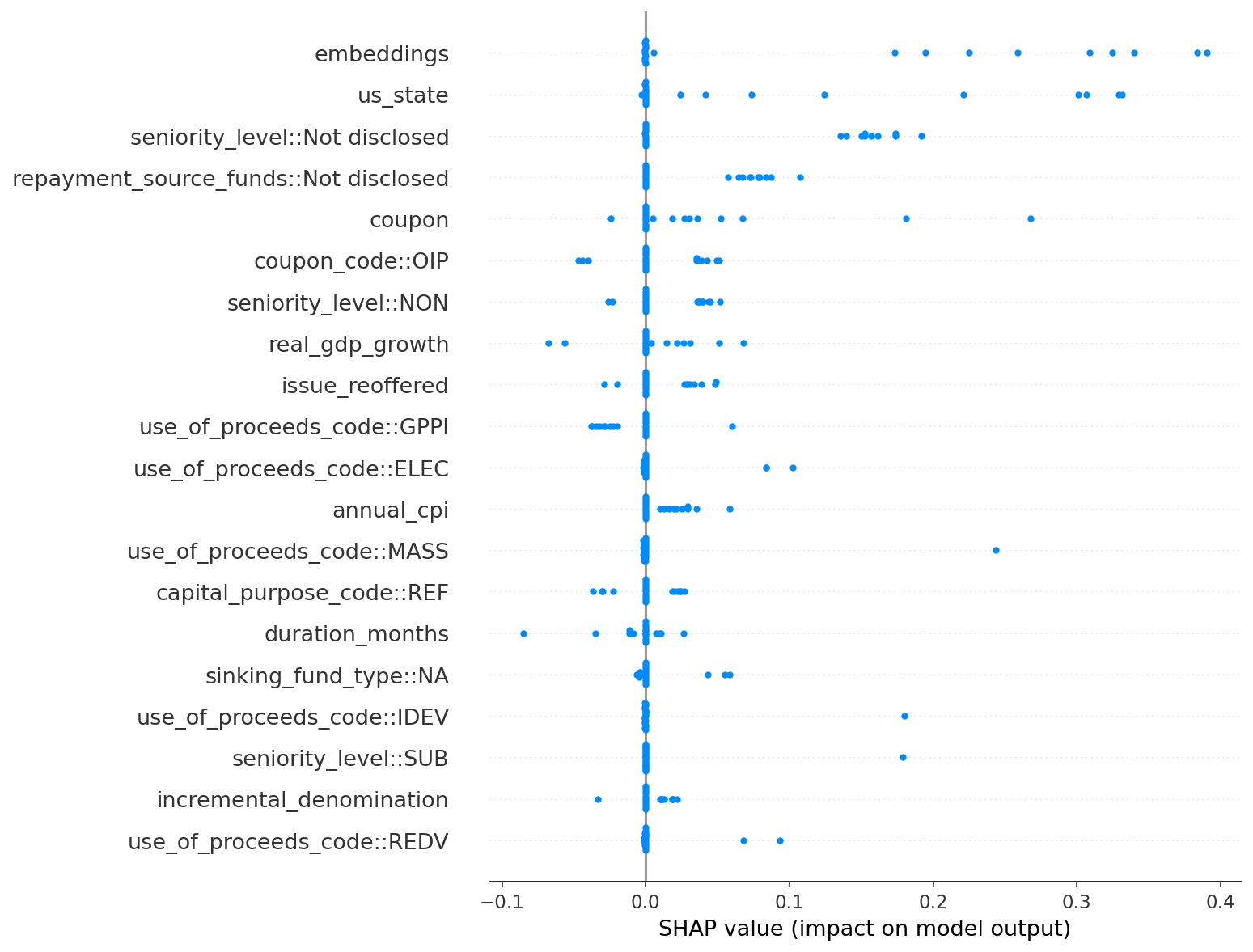}
  \caption{SHAP values for MLP model}
  \label{fig:shap}
\end{figure*}

To understand the performance of the neural network model, we plot the Shapley Additive Explanation (SHAP) values in Figure~\ref{fig:shap}, using the methods of \cite{shap2017paper}. It is clear that the dominant influences on the model's prediction are the embeddings of the project descriptions combined with states and the use of proceeds codes for the financing. A second cluster of features relates to the combination of macroeconomic conditions and the bonds' prior stated coupon and duration. A third cluster relates to more technical characteristics of the debt itself, such as the source of repayment funds and seniority level, with the model particularly penalizing non-disclosure on such terms (although still less than the embeddings of descriptions and geography).

In accord with this interpretation of the SHAP values, an ablative analysis shows that the synthetic oversampling is significantly less important for the MLP architecture than are the embeddings. That stands in contrast to the gradient-boosted ensemble methods, where synthetic oversampling is necessary to obtain a meaningful improvement over simple linear models. The difference seems due to the neural network's ability to learn to extract meaning from the interaction of the terms in the project descriptions (alongside the ability of batch-based training, with careful sampling, to undersample without losing too much information). 

These results have several implications. The first is that the use of transformer-based embeddings of loan project descriptions could be a fruitful avenue of research in other areas of credit. Such embeddings, which a great deal of research has by now shown to be superior to traditional embeddings in classification tasks in general, a result now confirmed in the specific domain of sub-sovereign public debt, could be applied in many other types of public and project-based credit issuance. The use of such embeddings may also, in such domains, finally allow the full exploitation of deep learning based methods, in domains where in prior work deep learning methods have struggled to meaningfully outperform ensemble-based methods.

A second implication is practical, and relates to the risk weightings of municipal bonds. Within our test set, a risk evaluation using the methods described here would have resulted in capital loss in less than 1 in 10,000 bonds. Any deployment of such a model for public credit decisions in real-life would have to be made with extreme care and a high ethical bar, notably on fairness and the effect on historically marginal geographies (though the order-of-magnitude lower false positive rate for the models compared to the yield spreads offered by human bidders provides some reassurance that such a bar could be met). If such concerns are addressed, the reduction in default rates through the use of models such as those presented here should motivate strongly for a further reduction in the risk weighting of municipal bonds, including revenue-based bonds. There would seem, for example, little means to justify assigning the same risk weighting (50\%) to residential mortgages and to revenue bonds, as the Basel III weightings now do, if default risk on municipal bonds were brought down to the levels described here. The large-scale reallocation of credit from the housing market to local public goods that might result from such a change would have substantial effects.

A third implication concerns future work. One avenue has already been implied: the extension of these methods to other forms of credit and jurisdiction. A second is the possibility of using these methods to predict other useful characteristics of municipal bonds, among them yields at offer. Another direction of future work could include building on the results here for reinforcement learning. That could have implications for bond trading, and hence improvements in municipal bond market making and liquidity. On the other hand, any such work will need to take into account the elevated trading costs in municipal bonds, with such trading costs being a known weakness in reinforcement learning based trading models.

Finally, we have here considered whether a municipal bond will fail to fulfill its \emph{financial} promises, but not its social and economic ones-that is, whether it will have the impact it promises. That task may be challenging, and will likely require the use of methods from growing literature on causal inference, and perhaps the ability to learn from relational context being developed in graph neural network methods. Nonetheless, even moderately reliable prediction of impact could have far-reaching practical implications, as well as further opportunities for transfer to other domains. In all, we believe that the careful combination of new AI methods to the large domain of funding local public goods has only begun to be explored, but already shows significant promise for important theoretical and practical advances.

\bibliographystyle{unsrt}
\bibliography{muni-bond-bib}

\end{document}